\documentclass{article}
\usepackage{spconf,amsmath,graphicx}
\usepackage{subfigure}
\usepackage{CJKutf8}

\title{Direct Speech-to-speech Translation without Textual Annotation using Bottleneck Features}
\name{Junhui Zhang, Junjie Pan, Xiang Yin, Zejun Ma}
\address{ByteDance AI-Lab}
\begin{document}
%
\maketitle
\begin{abstract}
Speech-to-speech translation directly translates a speech utterance to another between different languages, and has great potential in tasks such as simultaneous interpretation. State-of-art models usually contains an auxiliary module for phoneme sequences prediction, and this requires textual annotation of the training dataset. We propose a direct speech-to-speech translation model which can be trained without any textual annotation or content information. Instead of introducing an auxiliary phoneme prediction task in the model, we propose to use bottleneck features as intermediate training objectives for our model to ensure the translation performance of the system. Experiments on Mandarin-Cantonese speech translation demonstrate the feasibility of the proposed approach and the performance can match a cascaded system with respect of translation and synthesis qualities.
\end{abstract}
\section{Introduction}
\label{sec:intro}

Speech-to-speech translation (S2ST) tasks aim to translate a speech utterance from one language to another and have drawn increasing interests in recent research studies. The conventional solution of cascaded pipeline\cite{C2} usually includes three components, i.e., automatic speech recognition (ASR), machine translation (MT), and text-to-speech synthesis (TTS). This pipeline can produce satisfying results, but the performance may highly rely on the capability of each component, and error accumulation is a common problem for it. Typically, ASR, MT and TTS are all trained on text-dependent datasets and the required amounts and distributions of the datasets may vary significantly. As a result, constructing such pipeline is very expensive. Recent studies have proposed end-to-end (E2E) S2ST models \cite{C3,C4,C5} without relying on intermediate text representations. This is a much harder task but can effectively avoid error accumulation compared with the cascaded pipeline. For E2E S2ST models, training on a single parallel dataset, which is often very scarce, to achieve the performance of the cascaded system makes the task even challenging.

Translatotron \cite{C3,C5} proposes an end-to-end model with an auxiliary module to ensure the encoder outputs contain enough content information for phoneme sequences prediction from source and target speech. Dong \cite{C8} and Jia \cite{C9} introduce some data augmentation and finetune technics to help solve problems of data scarcity and further improve the performance. However, accomplishing such auxiliary task still requires speech transcripts and the performance of the entire network may highly rely on that of the auxiliary module. Speech-to-unit translation models \cite{C4,C7} are also proposed with self-supervised features like HuBERT \cite{C10} and discrete units \cite{C11} to accomplish this task. Lee \cite{C6} further proposes a textless S2ST to predict discrete units with a CTC finetuning technique for the removal of non-textual information such as speaker and accent in the discrete units.

Bottleneck features (BNF) have become widely used in TTS and voice conversion tasks \cite{C12,C13}. BNF are continuous frame-level features, which are the encoder layers’ outputs of the acoustic model in ASR. Since the acoustic model is trained to predict phoneme sequences with a supervised manner, BNF would contain more context and less acoustic information as they go deeper in the encoder layers, and BNF of the last few layers can be regarded as speaker-independent linguistic features with little acoustic information. These features can be regarded as naturally embedded content features similar to the embedded encoder outputs in MT tasks, and we find them potentially useful in S2ST tasks.

In this paper, we propose to use BNF as the intermediate supervised features to achieve a S2ST model which can be trained without any textual annotation. There is no auxiliary module for phoneme sequence prediction or non-content information removal, and the intermediate features can help model focus on the translation task in a supervised manner. Our experiments on Mandarin-Cantonese dataset show that this method can effectively translate between speech utterances from random source speakers with similar performance of the cascaded pipeline. The contributions of this paper are summarized as follows: 
\begin{itemize}
\item We propose a S2ST model trained without any textual annotation or auxiliary module for phoneme sequence prediction.
\item We introduce BNF with a supervised training strategy for model optimization.
\end{itemize}

In Section 2, we explain the proposed S2ST model in details. In Section 3, we present the subjective and objective experimental results to show the effectiveness of the proposed method. In Section 4, we summarize the progress we have made and discuss the future work.

\section{Methods}
\label{sec:method}

\subsection{Proposed system}
\label{proposed_system}

\begin{figure}
  \centering
  \includegraphics[width=6cm]{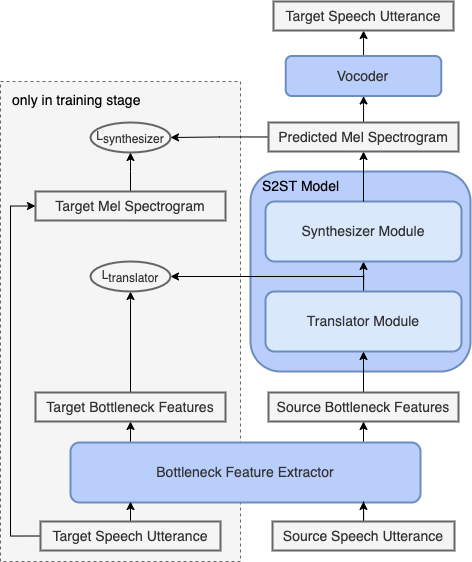}
  \caption{Proposed S2ST model pipeline.}
  \label{s2st_model}
\end{figure}

The proposed model along with the training and inference strategies are shown in Figure \ref{s2st_model}. The system contains a bottleneck feature extractor, a S2ST model and a neural vocoder. The bottleneck feature extractor is the encoder of an ASR model and outputs embedded features from its layers. The S2ST model receives the embedded features and outputs mel-spectrogram. The model consists of two submodules- a translator module and a synthesizer module. Details of these modules are explained in Section \ref{translator} and \ref{synthesizer}. A neural vocoder is used to transform mel-spectrogram to speech waveform. 

During the training stage, the extracted BNF are sent as inputs to the S2ST model. The translator module in the S2ST model predicts the intermediate embedded features $f_{pred}$ with a training target of BNF $f_{tgt}$ extracted from the target speech utterance. BNF from source and target speech can be considered as embedded content features, and the purpose of the translator is to translate between BNF of speech utterances. The synthesizer module, on the other hand, receives $f_{pred}$ to predict the mel-spectrogram $Y$ of the target speech. In the inference stage, BNF of source speech is sent into the S2ST model for mel-spectrogram prediction. Finally, a neural vocoder is used to generate waveform. This proposed method transforms speech utterances between different languages and from random source speakers to a target speaker.

\begin{figure}
  \centering
  \includegraphics[width=\linewidth]{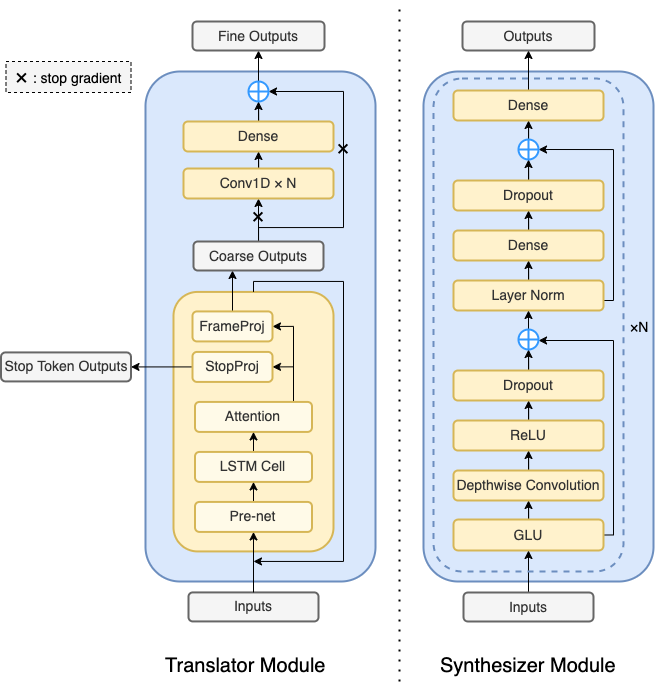}
  \caption{Detailed module structures in the proposed model.}
  \label{modules}
\end{figure}

\subsection{Translator module}
\label{translator}

The structure of the translator module is shown in the left part of Figure \ref{modules}, which is similar to a Tacotron \cite{C14} decoder. It takes BNF of source speech as inputs and auto-regressively outputs the intermediate features $f_{pred}$ and stop tokens $s_{pred}$. The auto-regressive module consists of a prenet with dense layers, a LSTM cell with uni-directional LSTM layers, an attention module and projection layers. We use GMM attention as in \cite{C15} to calculate and apply attention weights in the decoder. A frame projection layer and a stop-token projection layer predict coarse feature outputs $f_{c\_pred}$ and stop tokens $s_{pred}$, respectively. The coarse feature outputs then go through a post-net with a stack of convolution layers and a residual connection for the fine outputs $f_{pred}$ prediction. Stop gradient is used between the coarse outputs and the post-net. The training objectives of the coarse and fine outputs are both $f_{tgt}$, and the objectives of $s_{pred}$ is the ground-truth stop-token $s_{tgt}$. This helps the translator module learn to translate between source and target BNF in a supervised manner. The loss function for the translator module is the following.
\begin{equation}
    \begin{aligned}
     L_{translator} &= L_{coarse} + L_{fine} + L_{stop\_token} \\
                    &= \ell(f_{c\_pred}, f_{tgt}) + \ell(f_{pred}, f_{tgt})\\
                    & \ \ \ \ + \ell(s_{pred}, s_{tgt})
    \end{aligned}
  \label{loss1}
\end{equation}
where $\ell$ denotes L2 loss.

\subsection{Synthesizer module}
\label{synthesizer}

The synthesizer module uses a modified version of Light Convolution Blocks \cite{C16}. It contains a stack of blocks consisting of GLU, depth-wise convolution, dropout and normalizing layers with residual connections as shown in the right part of Figure \ref{modules}. During training, L2 loss is applied between every block output $y_i$ and target mel-spectrogram $Y$. To match the feature dimensions, each block outputs is followed by a dense layer to ensure $y_i$ has the same dimension as $Y$. The loss function for the synthesizer module is the following.
\begin{equation}
    \begin{aligned}
     L_{synthesizer} = \sum^N_{i=1}\ell(y_i -Y)^2
    \end{aligned}
  \label{loss2}
\end{equation}
where $\ell$ denotes L2 loss.

\subsection{Training objective and strategy}
\label{objective}

In the early training stage, the target BNF can be used directly as inputs of the synthesizer module for faster convergence of the two modules. And in the later training stage, two modules can be jointly trained for better performance and avoid error accumulation. Overall, the training objective of the proposed S2ST model is the following.
\begin{equation}
    \begin{aligned}
     L &= L_{translator} + L_{synthesizer} \\
       &= \ell(f_{c\_pred}, f_{tgt}) + \ell(f_{pred}, f_{tgt})\\
       & \ \ \ \ + \ell(s_{pred}, s_{tgt}) + \sum^N_{i=1}\ell(y_i -Y)^2
    \end{aligned}
  \label{loss3}
\end{equation}
where $\ell$ denotes L2 loss.

\section{Experiments}
\label{sec:exps}

\subsection{Datasets}
\label{dataset}
We conduct experiments on Mandarin-Cantonese language pairs, which can be considered as relatively distant languages in the same language family. Although they both use Chinese characters, their written forms differ significantly in grammar, idioms, word usage, etc., and the character pronunciations are also different because they have different phoneme sets. For example, ``\begin{CJK}{UTF8}{gkai}你不要开玩笑\end{CJK}" (Don't joke) in Mandarin would be expressed as ``\begin{CJK}{UTF8}{gkai}你\begin{CJK}{UTF8}{bkai}冇\end{CJK}讲笑\end{CJK}" in Cantonese, which have different expressions and the speech have totally different pronunciations. As a result, this language pair is representative in speech-to-speech translation tasks and is used in our experiments. Due to the scarcity of such parallel speech translation dataset, we generate it with a MT and a TTS model. Taking the advantages of existing multi-speaker TTS models and a trained Mandarin-Cantonese MT model following \cite{C17}, we translate the collected Mandarin scripts into Cantonese and synthesize the speech utterances with the TTS models. Multi-speaker TTS models are used to synthesize Mandarin speech with different speakers, and a single TTS model with Cantonese speaker is used to synthesize the Cantonese speech. In our experiments, we generate a parallel speech corpus with 1200hr speech pairs and experiment on different dataset sizes in training to observe the influence of dataset size to translation performance. Note that this approach is unnecessary if there is an existing parallel speech corpus, and no textual annotation is needed in training or inference stages.

\subsection{Implementation details}
\label{details}
The feature extractor is the encoder of a Conformer \cite{C18} ASR model and can generate BNF from different encoder layers. There are 36 layers in the encoder with feature dimensions of 512 and 40ms frame shifts, and we choose the 36th layer as the BNF in our experiments. Both the source and target speech utterances are extracted for S2ST model training.

For the proposed S2ST model, Table \ref{params} presents the detailed model structures of the translator and synthesizer described in Section \ref{translator} and \ref{synthesizer}. And we use a MelGAN\cite{C19} neural vocoder for the final waveform generation.

\begin{table}
\begin{center}
\begin{tabular}{ c|l } 
 \hline
  & Log-mel spectrogram of 24kHz wavs \\ 
 Spectral analysis & window size: 50ms \\ 
  & frame shift: 10ms \\ 
 \hline
  & prenet: FC-256 $\times$2 \\
  & LSTM cell: LSTMCell-256 $\times$2 \\
  & Attention: 8 GMM, hidden dim 128 \\
 Translator & Stop projection: FC-1 \\
  & Frame projection: FC-512 \\
  & Conv: kernel=5, channel=512, N = 5 \\
  & Dense: FC-512 \\
  \hline
  & GLU: kernel=3; filter=512 \\
  & Depthwise conv: kernel=17, strides=1 \\
 Synthesizer & Dropout: 0.1 \\
  & Dense: FC-512 \\
  & num blocks: N=6 \\
 \hline
\end{tabular}
\end{center}
 \caption{\label{params}Model configurations for the translator and synthesizer modules.}
\end{table}

\subsection{Results and analysis}
\label{results}
\subsubsection{Objective evaluation}
We conduct an objective analysis for the translation performance of the proposed model. Since there is no textual annotation or output in our model, in order to conduct BLEU score evaluation, we obtain the transcripts of source and generated speech utterances using a Mandarin and a Cantonese ASR model, respectively. The target translation text is generated using the Mandarin-Cantonese MT model as described in Section \ref{dataset} for BLEU calculation.

Table \ref{bleu} shows the BLEU results of the cascaded system and proposed method with different dataset sizes. The ground-truth represents the BLEU score of the MT model described in Section \ref{dataset}. The result shows that the model gets better performance as the size of parallel corpus increases, and an over 900-hour dataset is necessary for a good performance, which is close to that of a cascaded pipeline.

\begin{table}
\begin{center}
\begin{tabular}{ l|c } 
 \hline
 \multicolumn{1}{c|}{\textbf{System}} & \textbf{BLEU} \\
 \hline
 proposed model (300hr) & \ \ 1.81 \\ 
 proposed model (600hr) & 28.58 \\ 
 proposed model (900hr) & 33.53 \\ 
 proposed model (1200hr) & 40.84 \\
 \hline
 Cascaded pipeline & 41.60 \\
 Ground-truth & 65.66 \\
 \hline
\end{tabular}
\end{center}
 \caption{\label{bleu}BLEU results of the cascaded pipeline and proposed model. Duration of training corpus is noted in the brackets. The transcript is obtained by a Cantonese ASR model.}
\end{table}

\subsubsection{Subjective evaluation}
We split the subjective evaluation into content and speech level. 20 Cantonese linguistic experts are asked to give subjective scores to 40 translated speech utterances with random source speakers from the cascaded pipeline and our proposed model. The content level evaluation uses a set of carefully designed subjective criterion called Translation Mean Opinion Score (TMOS) described in Table \ref{tmos}. TMOS has the following advantages over BLEU. First, it can take grammatical mistakes into account instead of using only n-gram matching. Second, it focuses on sentence meanings so that the results no longer depend on the key words in target sentences. Besides, there is no fixed reference, so it can take different synonyms and expressions into account. These advantages make TMOS a reliable subjective evaluation method. For speech level, we simply use Mean Opinion Score (MOS) as the criteria. The linguistic experts are asked to score the quality of the synthesized speech with respect to naturalness and intelligibility.

\begin{table}
  \centering
  \begin{tabular}{l|c}
    \hline
    \multicolumn{1}{c|}{\textbf{Translation Result}} & \textbf{Score} \\
    \hline
    No output/More than half of the output is garbled     & 1.0 \\
    Output semantically irrelevant text                   & 2.0 \\
    Output semantically imperfectly relevant text         & 3.0 \\
    Error-prone direct translation                        & 4.0 \\
    Error-free direct translation                         & 5.0 \\
    \hline
  \end{tabular}
  \caption{\label{tmos}Subjective evaluation TOMS criteria.}
\end{table}

\begin{table}
\begin{center}
\begin{tabular}{ l|c|c } 
 \hline
 \multicolumn{1}{c|}{\textbf{System}} & \textbf{TMOS} & \textbf{MOS} \\ 
 \hline
 proposed model (300hr) & 0.93 & 1.81 \\
 proposed model (600hr) & 3.07 & 3.16 \\
 proposed model (900hr) & 3.26 & 3.25 \\
 proposed model (1200hr) & 3.85 & 3.87 \\
 \hline
 Cascaded pipeline & 4.07 & 4.14 \\
 \hline
\end{tabular}
\end{center}
 \caption{\label{subjective}Subjective TMOS and MOS results of the cascaded pipeline and proposed model.}
\end{table}

Table \ref{subjective} presents the subjective evaluation results of TMOS and MOS. Overall, the performance of proposed model has better performance as size of the training data increases. On content level, our proposed model achieves a TMOS close to the cascaded pipeline and is capable to translate normally with occasional mistakes. On speech level, MOS of proposed model is slightly lower than the cascaded pipeline, and the difference is mainly caused by some badly translated samples. The subjective evaluation results shows that the proposed model can achieve a close performance with the cascaded pipeline, and is potential with more datasets.

\begin{figure}
  \centering
  \includegraphics[width=\linewidth]{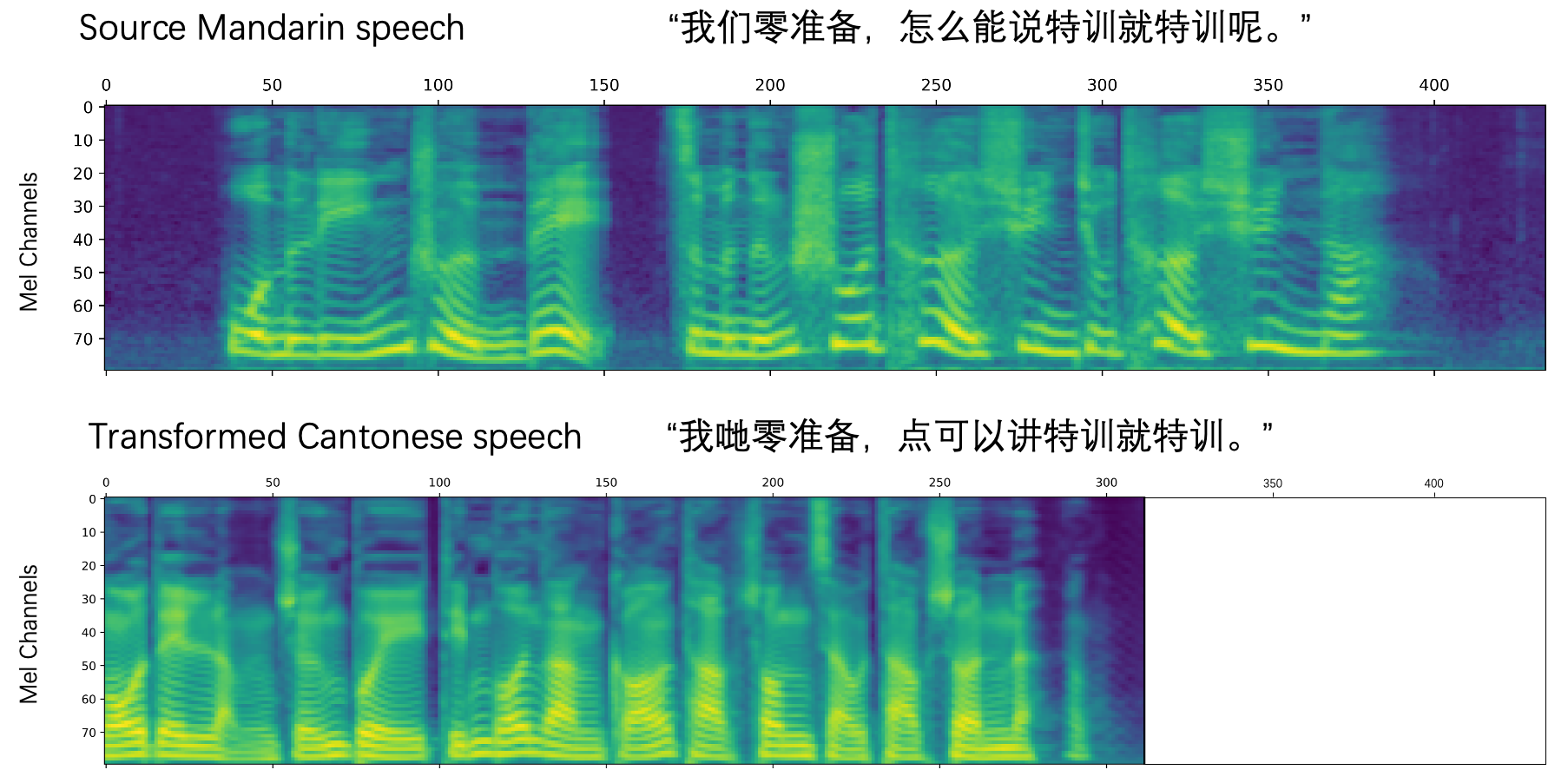}
  \caption{Mel-spectrograms of source and translated speech utterances.}
  \label{melspec}
\end{figure}

Figure \ref{melspec} presents the mel-spectrograms of a source Mandarin speech and a translated Cantonese speech. The model successfully transforms the speech from a female speaker to the target male speaker with different fundamental frequencies, and the durations of two utterances are different from each other corresponding to the lengths of text transcripted by linguistic experts. The speech demos are available online\footnote{demo page: https://zhangjh915.github.io/icassp-2023-s2st/}.

\section{Conclusions}
\label{sec:conclusion}

In this paper, we propose a direct speech-to-speech translation model without any textual annotation or content information using bottleneck features trained with a supervised manner. It accomplishes the task without auxiliary tasks of phoneme sequence prediction or non-textual information removal. The overall performance regarding translation and synthesis quality of proposed method is close to a cascaded pipeline and is potential with more datasets for training. For future work, we will adapt the model to different language pairs to evaluate its robustness, and can further design the synthesizer with a speaker module for multi-speaker translation.

\vfill\pagebreak

\bibliographystyle{IEEEbib}

\end{document}